\documentclass[11pt]{article}
\usepackage{amsfonts, amsmath, amssymb, amsthm, bbm, bm, color, enumitem, forloop, framed, graphicx, lscape, mathrsfs, pdfpages, rotating, sectsty, setspace, threeparttable, tocloft,multirow,verbatim,rotating,caption,subcaption}

\usepackage[title]{appendix}
\usepackage[backref=page]{hyperref}
\renewcommand*{\backref}[1]{}
\renewcommand*{\backrefalt}[4]{({\small%
    \ifcase #1 Not cited.%
          \or Cited on page~#2.%
          \else Cited on pages #2.%
    \fi%
    })}
\usepackage[T1]{fontenc}
\usepackage{microtype}

\usepackage{tikz}
\usetikzlibrary{matrix,chains,positioning,decorations.pathreplacing,arrows}

\usepackage[round]{natbib}  %add option -sort- to prevent multiple cites like (Author 2017c,b,a)

\hypersetup{pdfborder = {0 0 0},colorlinks=true,linkcolor=blue,citecolor=blue}

\numberwithin{equation}{section}

\theoremstyle{definition}
\newtheorem{remarkTmp}{Remark}

\newtheorem{egTmp}{Example}

\allowdisplaybreaks[1]

\setenumerate[1]{label=\bf(\arabic*)}
\setenumerate[2]{label=\bf(\roman*)}

\usepackage{marginnote} %for math mode margin notes.

    \def\ddefloop#1{\ifx\ddefloop#1\else\ddef{#1}\expandafter\ddefloop\fi}
    \def\ddef#1{\expandafter\def\csname c#1\endcsname{\ensuremath{\mathcal{#1}}}}
    \ddefloop ABCDEFGHIJKLMNOPQRSTUVWXYZ\ddefloop
    \def\ddef#1{\expandafter\def\csname s#1\endcsname{\ensuremath{\mathsf{#1}}}}
    \ddefloop ABCDEFGHIJKLMNOPQRSTUVWXYZ\ddefloop
	\def\ddef#1{\expandafter\def\csname b#1\endcsname{\ensuremath{\bm{#1}}}}
	\ddefloop aAbBcCgGhHjJMpPqQsStTwWxXzZyY\ddefloop
\begin{document}

\title{
	
	Foundation Priors \thanks{
         I would like to thank Max Farrell, Nabarun Deb, Chris Hansen, Alex Imas, Kevin Lee, Bradford Levy, Sendhil Mullainathan, Sridhar Narayanan, Oded Netzer,  Avner Strusov-Shlein, and Walter Zhang for helpful discussions and suggestions. 
         The paper has also benefited from feedback from discussions and presentations the Stanford AI and Marketing Conference, the Booth Marketing, Booth  Econometrics \& Statistics, and Wharton Marketing workshops. Numerous other conversations with industry experts at Google, Anthropic and Perplexity are also acknowledged. A number of language models including GPT (various versions), Claude, Gemini and Deepseek were used for research assistance. All written content is my own and I am responsible for all errors.   
        %at \url{asdf}.
        
    }
}

\author{
	Sanjog Misra\thanks{Booth School of Business, University of Chicago} 
}

%or use package authblk like this
%\author[1]{Max H. Farrell}
%\author[2]{Tengyuan Liang}
%\author[2]{Sanjog Misra}
%\affil[1]{University of California at Santa Barbara}
%\affil[2]{Chicago Booth School of Business}

\date{}

\maketitle

\begin{abstract} 

Foundation models, and in particular large language models, can generate highly informative responses, prompting growing interest in using these “synthetic” outputs as data in empirical research and decision-making. This paper introduces the idea of a \emph{foundation prior}, which shows that model-generated outputs are not as real observations, but draws from the foundation prior induced prior predictive distribution. As such synthetic data reflects both the model’s learned patterns and the user’s subjective priors, expectations, and biases. We model the subjectivity of the generative process by making explicit the dependence of synthetic outputs on the user’s anticipated data distribution, the prompt-engineering process, and the trust placed in the foundation model. 

We derive the foundation prior as an exponential-tilted, generalized Bayesian update of the user’s primitive prior, where a trust parameter $\lambda$ governs the weight assigned to synthetic data. We then show how synthetic data and the associated foundation prior can be incorporated into standard statistical and econometric workflows, and discuss their use in applications such as refining complex models, informing latent constructs, guiding experimental design, and augmenting random-coefficient and partially linear specifications. By treating generative outputs as structured, explicitly subjective priors rather than as empirical observations, the framework offers a principled way to harness foundation models in empirical work while avoiding the conflation of synthetic “facts” with real data.

\end{abstract}

\bigskip


\setcounter{page}{0}
\thispagestyle{empty}

\newpage
%\onehalfspacing
\doublespacing

\section{Introduction}

%
%\begin{itemize}
%    \item Focus on subjectivity
%    \item Contexts cumulate
%%    \item Heterogeneity
 %   \item Sources: Hidden Ability and Action
%    \item f(q1,q2) --> f_1(q2)
%    \item redo as sampler rather than SGD?
%    \item Prior strength matters: 1+1 vs something new/more % abstract
%    \item Why it matters - agents. 
%    \item example: from ML school
    
% \end{itemize}
%\textbf{
%\begin{itemize}
%    \item f-prior can be on any %part of $\theta$
%    \item so random coefficients
%\end{itemize}}

Foundation models such as large language models (LLMs) now generate text and other content that can appear remarkably informative. Every output, however, is the product of an interaction between the model’s learned statistical patterns and the user’s prompt. We posit that this production process renders outputs from  foundation models inherently subjective. Since LLMs are used to guide research, policy, and business choices, their epistemic role is defined both by their informativeness and the subjectivity embedded in their generation. A subtle shift in the meaning of “data” is taking place: Knowledge once derived from empirical observation is now supplemented, or in some cases replaced, by information that is co-produced through human–model interaction. In this sense, generative outputs function as a new form of subjective data, blending learned correlations with user-imposed priors and serving as inputs to analysis, training, and belief revision.\footnote{Throughout, we use the term “generative data” to denote responses from foundation models. This terminology emphasizes their data-like use and clarifies the issues under discussion. The terms “generative” and “synthetic” data are used interchangeably.}

A key concern with such generative data is that its provenance is uncertain. Put differently, the process that generates the data is not cleanly delineated. This makes the reliability of such data for statistical inference uncertain. For one, we often have minimal visibility into the structure of foundation models and limited knowledge of the precise data used to train them. Moreover, as we will show, the prompt design and engineering process injects the user's own subjective priors, beliefs, and preferences into the generation mechanism. This complex and often opaque generation process raises questions about the nature of the generated data and concerns about treating them as equivalent to objectively collected empirical or "real" data. 

This paper's central thesis is that such generative data should primarily be interpreted as samples drawn from a specific type of \textit{subjective} prior predictive distribution. We term this prior the \emph{foundation prior} - a generative scheme that synthesizes responses to user queries in a way analogous to drawing from a parameterized distribution but one that is intractable and subjectively malleable. Our aim is to offer a structured path forward: by treating generated outputs as an explicit form of prior knowledge, one can harness their richness while avoiding the risks of conflating synthetic “facts” with empirically grounded evidence.

We  begin by formalizing how a user, equipped with a preexisting prior over the construct of interest, forms an anticipated data distribution, essentially, what they expect generated data to look like. We then model prompt engineering as an iterative alignment process where the user proposes a query to the foundation model, evaluates the resulting synthetic data against the anticipated distribution (using a divergence measure) and refines the prompt until the synthetic data aligns sufficiently with those priors. The end product is a generative, synthetic dataset that captures both the foundation model’s learned patterns and the user’s subjective filters. We then demonstrate how one might incorporate this dataset into a decision framework, tempering it by a “trust” parameter $\lambda$ that determines how heavily to lean on synthetic information versus the original prior. The result is the foundation prior — a structured distribution reflecting the user’s beliefs, the generative model’s knowledge, and the subjective alignment processes required to reconcile them. This construction offers a principled way to employ synthetic model outputs in empirical analysis without conflating them with raw observations or data from the real world.

In part, this paper is motivated by current practices, both in academia, as well as, in industry\footnote{A large number of firms are exploring the creation and use of synthetic data and respondents. For example, in the consumer data realm we have firms like NielsenIQ, Kantar and IRI among others who have begun to explore the use of such synthetic data. There are also companies like \href{https://www.syntheticusers.com/}{syntheticusers.com} providing survey based responses based on generative models and have taglines such as "User Research. Without the users."}, that seek to use data generated by foundation models in some form of statistical or econometric analysis. On the academic front, the contexts in which such ideas emerge are spread across a variety of disciplines and applications in the medical and social sciences including economics, business, psychology, political science, among others. In particular, there is now a growing literature that speaks to the ability of foundation models to simulate human behavior in contexts such as survey response (\cite{park_generative_2024}), psychological tests and experiments (\cite{dillion_can_2023}), political behavior (\cite{argyle_out_2023}) or economic interventions (\cite{horton_large_2023}). Many of these papers propose, discuss, or evaluate the potential for generative, synthetic data to replace human-generated data. The general tone of the emerging literature pertaining to the use of synthetic data is cautiously optimistic, with some concerns about issues pertaining to bias, fidelity to reality and measurement error issues. 

A substantial literature treats models trained on historical data as objective priors for new inference problems. Classical Bayesian work develops power priors \citep{ibrahim_power_2000}, expected-posterior priors \citep{perez_expected-posterior_2002}, intrinsic and catalytic priors \citep{berger_intrinsic_1996}, and commensurate or meta-analytic priors for borrowing strength across studies \citep{hobbs_commensurate_2012}. These approaches share the perspective that historical data, or a model summarizing such data, can be incorporated as a fixed, external source of prior information. This logic also appears in empirical Bayes formulations \citep{carlin_bayes_2000}, where past data are treated as providing an estimate of a baseline distribution that is then used as a prior for new likelihoods. This broad idea is adapted by \citep{ohagan_ai-powered_2025} who propose procedures that use foundation models as a source of objective prior information information. 

Modern machine-learning approaches extend this idea by treating complex generative models as implicit priors. Prior-data fitted networks, including TabPFN and its variants \citep{hollmann_accurate_2025}, train foundation-style models entirely on synthetic data generated from a known prior so that the resulting transformer behaves as an amortized Bayesian update. Extensions to causal inference, such as CausalFM \citep{ma_foundation_2025}, follow the same principle by generating synthetic data from structural causal models and using these draws as the prior-data regime for pretraining. In applied domains, generative models have been used as high-dimensional priors in clinical trial design \citep{arai_generative_2025}, seismic inversion \citep{xie_generative_2024}, and Bayesian evidence computation in astrophysics \citep{mcewen_bayesian_2024}. Across these literatures, the fundamental assumption is that the historical-data model, whether statistical, empirical-Bayes, or foundation model based, acts as an \emph{objective} prior that is fixed with respect to the analyst.

Emerging literature also delves into the possibility of using these responses as data in non-Bayesian approaches. \cite{ludwig_large_2025} develop an econometric framework that classifies LLM applications into prediction and estimation tasks. They demonstrate that LLM outputs can be useful for hypothesis generation and predictive modeling, but only under the strict condition of no training leakage, meaning that the generative model must not have been trained on the same data as the research sample. For estimation problems, the authors argue that researchers must collect validation data to account for the potential bias introduced by measurement error.  

Our work differs from the extant literate in that we explicity acknowledge and model the subjectivity inherent in generated, synthetic data. As such, our approach complements the literature by considering the conditions under which generative models serve not just as useful in predictive tasks but as structured priors that can be used for empirical inference. In doing so, we offer an alternate, coherent framework that acknowledges the potential for epistemic circularity and distributional mismatch yet offers avenues for numerous use cases and applications.

We, instead, focus on the complexity inherent in the \emph{generation} of such data and the issue of subjectivity introduced by the user in the generative framework. Our analysis provides a coherent framework that allows the interpretation of the output of foundation models as those conditioned on a complex prior, and offers a guide to appropriate avenues for the use of such data.   In what follows, we model the generative process that the user follows to generate data, characterize the foundation prior and following that delve into the use of this prior and synthetic data for analysis and decision making.

\section{The Foundation Prior}

\subsection{Generative Process}

We start with describing the generative process. Typically a user will provide a prompt or query ($q$) which then generates some data ($D_s$) from the model which is then used to learn about some construct ($\theta$). We will describe and model two broad components for the interaction the user has with the generative model. First, we describe the process of anticipation - that is, the user has expectations over the type of data or response they will accept from the generative process. Second, we argue that that users will reject data that does not conform to these anticipations and iterate on refining the prompt until they find the outcome acceptable.    

\subsection{Anticipation}

Before the user starts the generative process, they anticipate the nature of the data to come. 
As an example, suppose that the user wishes to generate data about the sales of ice-cream in a particular city (say Chicago). They would \textit{anticipate} that that the generated would concur with their prior beliefs, say with summer sales being higher than those in winter.  

This process of anticipation involves forming beliefs about the data that a generative model will produce before actually querying the model. To form an anticipation of the data ($D_a$), the user constructs a data distribution based on their model specification. In other words, given the likelihood function $L(D | \theta)$, which models how data $D$ is generated given $\theta$, and priors $\pi(\theta)$, the marginal anticipated data distribution can be expressed as:
\begin{equation}
    \pi(D_a) = \int L(D_a | \theta) \pi(\theta) d\theta.
\end{equation}
This integral represents the marginalization over all possible parameter values under the prior distribution. It provides a measure of what data one should expect a priori, before engaging with the generative model. We note here that the user is self-consistent in the likelihood $L$ reflects the user's beliefs about the true data generating process. It is, consequently, the same likelihood that the user would use to model real data when available. If $\pi(\theta)$ is informative and strongly concentrated around a specific region, the anticipated data distribution $\pi(D_a)$ will be narrow, indicating very specific anticipated data. Conversely, a diffuse prior results in a broader anticipated distribution, reflecting greater uncertainty.

\subsection{Prompt Engineering}

We consider a process where a user refines the prompt $q$ iteratively, to adjust the output distribution of a language model (LLM). The user draws a dataset $D_s$ from the LLM and evaluates some divergence metric:
\begin{equation}
    \kappa(\pi(D_a) \| D_s(q) ),
\end{equation}
which measures the discrepancy between a anticipated data distribution $\pi(D_a)$ and the LLM's output  conditioned on $q$.\footnote{Note here that $\kappa$ can take a distribution as its first argument and either a distribution or an empirical dataset as its second. Examples for the latter include a negative log-likelihood or sample based MMD measures. In the sequel, we will use the dataset notation.}  Note that we do not assume that the user knows the generating process for $D_s$. We think of $\kappa$ as some distance or divergence  that the user constructs. For example, it could be a distance metric based on moment matches, a likelihood ratio, or more sophisticated constructs such as integral probability metrics. For the purposes of our analysis it suffices that the user seeks to match the generated data to their anticipation.  

The user updates $q$  with the aim to lower the discrepancy ($\kappa$) and stops when a stopping criterion is met. Below, we outline an abstract SGD-type\footnote{We remind the reader that this SGD process is an abstraction. We could use a rejection sampling approach, MCMC routine or some other iterative approach that accepts synthetic data based on its similarity to anticipated data.} policy for updating $q$ without assuming any particular parametrization and discuss possible stopping rules.

The user updates $q$ iteratively to refine the output distribution. At each iteration, the user draws a dataset $D_s(q_t)$ from the foundation model (LLM):
\begin{equation}
    D_s(q_t) \sim \pi_{\text{LLM}}(D_s \mid q_t),
\end{equation}
and computes $\kappa$. The update rule for $q$ is then:
\begin{equation}
    q_{t+1} = q_t - \eta_t \cdot \nabla_q \kappa(\pi(D_a) \| D_s(q) ),
\end{equation}
where $q_t$ is the prompt at iteration $t$, $\eta_t > 0$ is the learning rate and $\nabla_q \kappa$ is an estimated gradient of the divergence with respect to $q$. 

This is a stylized model of prompt engineering and one could add more nuance and structure to the model, for example,  by allowing the anticipated data distribution to evolve with the observation of synthetic data. The key point here is to acknowledge that the user's prompt at any iteration ($q_t$) depends on the anticipated data distribution ($\pi(D_a)$) and through it the user's prior $\pi(\theta)$.  As such, any generated data will also depend on those subjective elements. 

\subsubsection{Stopping Rules}

The user stops updating $q$ when a predefined condition is satisfied. There are several stopping rules that the user could adopt:
A natural stopping rule is when $\kappa$ falls below a threshold $\tau$:
\begin{equation}
   \kappa(\pi(D_a) \| D_s(q_t) ) < \tau.
\end{equation}
This ensures that the prompt produces an output distribution sufficiently close to the target. Conceptually, this rule implies that the synthetic data is similar to the anticipated data of the user. 

A second possibility is that the user stops updates when changes in $\kappa$ are small over consecutive iterations:
\begin{equation}
    |\kappa(\pi(D_a) \| D_s(q_t) )  - \kappa(\pi(D_a) \| D_s(q_{t-1}) ) | < \epsilon,
\end{equation}
where $\epsilon$ is a small tolerance value. This could happen if the prompt engineering process fails to reflect any changes in the divergence. As such, the user might give up on further attempts to match the synthetic data to their anticipated data distribution. 

Finally, the user could stops after a fixed number of iterations $T_{\max}$:
\begin{equation}
    t \geq T_{\max}.
\end{equation}
This rule probably emerges when there are information processing constraints present. In effect, a smaller number of iterations could also reflect the user's lower reliance on their prior and their willingness to accept a broader range of data generated from the language or foundation model.

The above discussion only seeks to outline an abstract iterative policy for refining prompts. Obviously, the actual prompt engineering process could be much simpler or more complex. What should be clear from the discussion though are (a) That the user's prior plays a significant role in the prompt engineering process via the anticipation mechanism, (b) That the prompt engineering process requires the subjective specification of various constructs such as the learning rate, the divergence and other relevant parameters, and (c) the stopping rule for the prompt engineering process reflects various subjective preferences or constraints on the part of the user. 

\subsection{Fine Tuning}

The anticipated data distributions can also be refined by incorporating additional information or domain constraints. If additional information $V$ is available, the posterior-adjusted anticipated data distribution is:
\begin{equation}
    \pi(D_a | V) = \int L(D_a | \theta) \pi(\theta | V) d\theta,
\end{equation}
where $\pi(\theta | V) \propto G(V | \theta) \pi(\theta)$ represents the posterior over parameters given observations $V$ for some model $G$. 

One example of this kind of data-augmented updating process is fine-tuning (\cite{howard_universal_2018}), where the foundation model parameters are adjusted based on some task specific auxiliary data ($D_\text{aux}$). Here, we can write the tuned model as $\widetilde{p}_\text{LLM}(D|q,D_\text{aux})$, where the change in notation ($p$ to $\widetilde{p}$ ) reflects the fact that the parameters of the original model have been altered. One might also reasonably expect that the user's on prior has been updated with the auxiliary data such that
\begin{equation}
    \pi(D_a | D_\text{aux}) = \int L(D_a | \theta) \pi(\theta | D_\text{aux}) d\theta.
\end{equation}
Now both the prior and the generative model have changed because of the fine tuning. In this case, the divergence can be written as    
\begin{equation}
    \kappa(\pi(D_a|D_\text{aux}) \| D_s(q,D_\text{aux}) ).
\end{equation}
The prompt engineering process now sits atop these altered beliefs. We reiterate that the ideas described above are  illustrative of what is, in all likelihood, a more complicated process. The key point is that the generative framework necessitates input from the user at various points and consequently injects considerable subjectivity into the process.   

\subsubsection{Interpreting Generative Data}
The prompt engineering process described above injects the user's assumptions,  priors and subjective beliefs into the generated data via some rejection process. The terminal prompt encapsulates this subjectivity and is used to generate acceptable data - we denote this prompt as $q^*$ and the corresponding data it generates as $D^*_s$. Based on our description above it should be clear that the generated data cannot be thought of as being the same as "real" data. There are a number of points worth emphasizing to make the distinction between real and generative data clear. 

First, what should be clear is that $D^*_s$ is based on a prompt $q^*$ which is a complex function of the user's prior and the anticipation process. The prompt engineering process enforces a form of selection. Since the user seeks to match the synthesized data to anticipated data they, in effect, reject data that does not match their prior beliefs. As such, $D^*_s$ is not simply drawn from the foundation model but is convolved with the users anticipation in a non-trivial manner. Moreover, since the anticipation process depends on prior beliefs of $\theta$ there is a serious issue of epistemic circularity. As such, $D^*_s$ while potentially informative, is also contaminated by the user's subjective beliefs. 

As we have shown, the prompt engineering process is a function of a number of subjective choices made by the user. We have already mentioned that it is influenced by the anticipation process and via it on $\pi(\theta)$. Further, choices pertaining to the learning rate ($\eta$), discrepancy measure ($\kappa$), the stopping rule and other elements of the prompt engineering process all play a role in the construction of $q^*$ and consequently on $D^*_s$. While our description of the generative process is stylized,  it reiterates the fact that the data emerging from language models are not objective.    

Moreover, the foundation model's internal generative process is based on learned representations and statistical approximations from its training corpus. In contrast, real data is generated by humans via some natural, more complex, underlying processes. For example, real data is often subject to unpredictable environmental influences, measurement errors, and other stochastic factors. In contrast, synthetic generated is generated in a controlled computational environment where such sources of randomness or error are either absent or not accurately modeled. This absence can make the synthetic data less representative of the complexities inherent in real data. A related issue is that of scale. Data generated from foundation models are conceptually infinite and since the cost of generation are virtually zero there is not particular reason to limit the amount of generative data used for analysis. As such, confidence measures and standard errors are meaningless and we would need to accept any inference for $\theta$ based on $D_s$ as being perfect. As a consequence, the need for real data is obviated which seems unreasonable. Even so, there is increasing evidence that the data generated by foundation models is informative and we need to appropriate the information contained in them into our analysis.

One might argue that any research design that elicits data from humans also encodes such subjectivity. For example, the wording of survey questions are subjective, as are model specifications or choices pertaining to sampling procedures and design. While we agree that this claim has merit, we would argue that the fundamental difference lies the fact that the prompt design and engineering process operates via a rejection mechanism. The prompt engineering process systematically rejects data that do not conform to the user's anticipation. If this were done with a real research design we might find the practice unpalatable.  Examples of such practices include p-hacking, data dredging and snooping, or in a more extreme case altering the data itself. Each of these effectively rejects the original data for a modified more acceptable version. We concede that there are also a number of less problematic procedures used in the real data collection context that might be considered similar to the rejection based prompt engineering process we described above. For example, running pilots (for an experiment) to see if the design delivers the type of data we need. We contend that these two can be thought of injecting subjectivity into the process but in a manner that the research community finds acceptable. In contrast, we conjecture, that users and the broader research community may not see the prompting process as one that injects subjectivity into the process and hope this paper alerts them to that issue. 

In summary, the generative process delivers us data that is potentially informative but convolved with the user's preferences and subjective choices. In what follows we discuss how we might use this data to construct beliefs about $\theta$.

\subsection{Foundation Prior: Characterization}
\label{subsec:foundation_prior_derivation}

Thus far we have articulated a descriptive mechanism that shows how the user tunes prompts to produce an acceptable synthetic dataset $D_s^*$. We now turn our attention to the characterization of beliefs about $\theta$ conditional on these data. We will take the perspective of a decision theorist: Given synthetic data $ D_s^*$ drawn from $\pi_{\text{LLM}}(\cdot \mid q^*) $, we aim to find a function $ \rho^*(\theta) $ that satisfies the following two objectives:

\begin{enumerate}
    \item Minimize deviation from the prior $ \pi(\theta) $ using the Kullback-Leibler (KL) divergence\footnote{We use the KL divergence for analytical tractability. Other divergence metrics may also be used at the cost of such tractability.}:
    \begin{equation}
        \min_{\rho} \text{KL}(\rho \parallel \pi) = \int \rho(\theta) \log \frac{\rho(\theta)}{\pi(\theta)} d\theta.
    \end{equation}
        \item Ensure that the expected log-likelihood of the synthetic data under $ \rho $ meets a constraint:
    \begin{equation}
        \mathbb{E}_{\rho}\left[\log L(D_s^* \mid \theta)\right] \geq C(\kappa^*).
    \end{equation}
\end{enumerate}
The first condition requires that the beliefs about $\theta$ after observing the synthetic data are not too far away from the user's initial priors. The second condition requires that under the new beliefs the expected log-likelihood (evaluated using the synthetic data) be above some acceptable levels. This optimization setup has its roots in the seminal work by \cite{jaynes_information_1957} and \cite{kullback_information_1959} and later \cite{csiszar_i-divergence_1975} and \cite{donsker_asymptotic_1975}. These authors provide the basic elements of this framework. More recently the approach has been used in numerous fields such as physics, statistical mechanics and statistics among others to derive posterior beliefs. 

The requirements above form a constrained optimization problem where we seek the optimal posterior-like function $ \rho^*(\theta) $ that balances prior adherence and likelihood performance. The term $C(\kappa^*)$ reflects the fact that the inequality threshold may be a function of the divergence $\kappa^*$ induced under the prompt $q^*$. Note that we are taking the synthesized data as given. Even though $D_s^*$ depends on $q^*$ and via it on $\pi(\theta)$ this influence is already encoded in the synthetic data and we do not need to address it.

To incorporate the likelihood constraint, we introduce a Lagrange multiplier $ \lambda \geq 0 $ and define the following Lagrangian:

\begin{equation}
\label{langrangian}
\mathcal{L}[\rho] = \underbrace{\int \rho(\theta) \log \frac{\rho(\theta)}{\pi(\theta)} d\theta}_{\text{KL term}} - \lambda \left( \underbrace{\int \rho(\theta) \log L(D_s^* \mid \theta) d\theta - C(\kappa^*)}_{\text{Constraint}} \right).
\end{equation}

Taking the functional derivative with respect to \( \rho(\theta) \):
\begin{equation}
\delta \mathcal{L} = \log \rho(\theta) - \log \pi(\theta) + 1 - \lambda \log L(D_s^* \mid \theta) = 0.
\end{equation}
Solving for the foundation prior \( \rho(\theta) \) we get:

\begin{equation}
\label{fprior}
\boxed{
\rho(\theta|D_s^* ,\lambda) = \frac{\pi(\theta) L(D_s^* \mid \theta)^\lambda}{\int \pi(\theta) L(D_s^* \mid \theta)^\lambda d\theta}.
}
\end{equation}

The foundation prior structure in \ref{fprior} is simple and intuitive. The form conforms to Bayes rule apart from the parameter $\lambda$. Indeed, as mentioned earlier, the works of \cite{csiszar_i-divergence_1975} and \cite{kullback_information_1959}, among others, formalize the idea of Bayes' theorem as an outcome of an optimization process. Our result also shares a structural similarity to the generalized Bayes updates of \cite{bissiri_general_2016}. The key difference is that in those papers the data is real while ours is generative. As a consequence the interpretation of $\lambda$ is also somewhat different. While the extant literature thinks of $\lambda$ as a robustness parameter in our context it aligns more with a trust interpretation. If $\lambda=0$ then the foundation prior reduces to the user's primitive prior $\pi(\theta)$ while if $\lambda=1$ then it suggests that the user trusts the synthetic data as much as they would trust anticipated data or for that matter real data. Note here that because the synthetic data likelihood sums over $N_s$ observations, its contributions to the foundation prior scales with $\lambda N_s$. As such, it is this product that establishes the effective strength of the prior. We will return to this issue shortly.

In the above analysis (and in most of what follows) we focus on a generic parameter $\theta$ that is of interest and can be informed by the generative outputs of a foundation model. Clearly, we could partition $\theta=\{\theta_1,\theta_2\}$ and use the generated data to inform only some chosen subset of $\theta$, say $\theta_1$. In some applications we discuss later we demonstrate the power of this idea of using synthetic data in specified parametric or structural models.

\subsection{Representation}

There are multiple representations of the foundation prior. In our discussions above we have described the prior formally via our specification of $\pi(\theta|D_s^*,\lambda)$. However, the information contained therein can be transmitted via alternative mechanisms and forms. We outline some of these here. 

Consider, for example, $D_s^*$. This generated data is based on a tailored $q^*$ which is designed to make the data match the distribution of anticipated data. As such, it can be thought of as an approximate draw from the foundation prior predictive density. More specifically, since $q^*$ encodes the user's priors and preferences into the generative process, the generated data will propagate those constructs. Obviously, repeated draws will give us multiple synthetic datasets, each of which can be used to estimate $\theta$ and thereby construct an alternative representation for the foundation prior.  

Yet another representation might be to ask the foundation model to simply generate \emph{code} that can provide draws of $\theta$ (or equivalently $D_s$) that can be used in subsequent analysis. We do not endorsing this approach, rather using it as a illustration of the point that the foundation prior can be represented in a variety of ways. While these representations are not technically equivalent, each of them communicates forward the information contained in the foundation prior.

\section{Addressing Subjectivity}

The foundation prior is inherently subjective because it depends on how an analyst interacts with a foundation model. Different analysts will phrase instructions differently, choose different examples, emphasize different aspects of the task, or use different prompting styles. Let $q$ denote the prompt or interaction specification used to generate synthetic data $D_s^*(q)$ from the foundation model. Because prompts vary systematically across individuals, the appropriate conceptual framework is not uncertainty about a correct prompt, but heterogeneity across the prompts people might naturally generate.

This heterogeneity propagates into the foundation prior
\[
\rho(\theta \mid D_s^*(q),\lambda)
= 
\frac{\pi(\theta)\,L(D_s^*(q)\mid\theta)^{\lambda}}
     {\int \pi(\theta)L(D_s^*(q)\mid\theta)^{\lambda}\,d\theta},
\tag{\ref{fprior}}
\]
where $\lambda$ is a trust parameter that determines the strength of the synthetic-data contribution relative to the user's primitive prior $\pi(\theta)$. Because $\lambda=0$ yields the primitive prior and $\lambda=1$ treats the synthetic data as fully credible, mitigating subjectivity requires tools that operate both across heterogeneous prompts $q$ and across different values of $\lambda$. We discuss three such tools: integration over heterogeneous prompts, robustness through conservative trust, and calibration using real data.

\subsection{Integration Across Heterogeneous Analysts}

The first strategy treats prompts as heterogeneous expressions of how different analysts would articulate the same data-generating question. Let $q$ index a heterogeneous prompt regime. For each $q$, the foundation model produces a synthetic dataset $D_s^*(q)$ and an associated foundation prior $\rho(\theta \mid D_s^*(q),\lambda)$. These variants reflect heterogeneity across analysts, not randomness. A natural way to incorporate this heterogeneity is to integrate across a distribution $h(q)$ representing the population of prompts individuals might reasonably produce:
\[
\bar{\rho}(\theta \mid \lambda)
=
\int \rho(\theta \mid D_s^*(q),\lambda)\,h(q)\,dq.
\]
The distribution $h(q)$ can be constructed empirically by eliciting prompts from a group of individuals or by using a structured family of prompts capturing realistic stylistic, semantic, or logical variations. This approach replaces dependence on a single idiosyncratic prompt with a population-average prior reflecting the diversity of human interactions with the model. The resulting prior $\bar{\rho}$ is therefore less brittle and more democratically grounded.

\subsection{Robustness Through Conservative Trust}

The parameter $\lambda$ already appears in the foundation prior as a trust parameter. In our setting, robustness is not about guarding against prompt misspecification (there is no single optimal prompt) but about guarding against the fact that \emph{another reasonable analyst} might have generated a different synthetic dataset by using a different prompt $q$. Lowering $\lambda$ therefore provides a principled mechanism for being more conservative in the face of heterogeneous analyst inputs.

Concretely, the likelihood contribution of synthetic data generated under prompt $q$ is $L(D_s^*(q)\mid\theta)$. The foundation prior assigns it weight $\lambda$ through the exponential tilt $L(D_s^*(q)\mid\theta)^{\lambda}$. Smaller $\lambda$ values downweight the synthetic component, reducing the influence of any single $q$ in cases where analyst heterogeneity is substantial. This parallels the classical Bayesian robustness literature, where downweighting protects against sensitivity to the prior. Here, the same logic protects against sensitivity to heterogeneous prompting behavior. If integration across $q$ averages over heterogeneity, robustness tempers it.

\subsection{Calibration Using Real Data}

A third mechanism for managing subjectivity is calibration: real data can be used to determine how much trust should be placed in the synthetic-data component. While the foundation prior reflects subjective information encoded in prompts, the real data carry objective information that can anchor or correct subjective variation.

Suppose real data $D_r$ is observed. Given synthetic data $D_s^*(q)$ generated under prompt $q$, the combined posterior is
\[
\pi_{sr,\lambda}(\theta)
\propto
\pi(\theta)\,L(D_s^*(q)\mid\theta)^\lambda\,L(D_r\mid\theta).
\]
The Donsker–Varadhan representation implies a variational identity for the marginal likelihood of $D_r$ under the foundation prior:
\[
\log m_\lambda(D_r)
=
\sup_{\rho}
\left\{
\mathbb{E}_\rho[\log L(D_r\mid\theta)]
-
\mathrm{KL}(\rho \,\|\, \rho(\theta\mid D_s^*(q),\lambda))
\right\},
\]
whose maximizer is $\rho^*(\theta)=\pi_{sr,\lambda}(\theta)$. Differentiating with respect to $\lambda$ gives
\[
\frac{\partial}{\partial\lambda}\log m_\lambda(D_r)
=
\mathbb{E}_{\pi_{sr,\lambda}}[\log L(D_s^*(q)\mid\theta)]
-
\mathbb{E}_{\rho(\cdot\mid D_s^*(q),\lambda)}[\log L(D_s^*(q)\mid\theta)].
\]
A natural calibration criterion is to choose $\lambda=\lambda^*$ such that the marginal value of increasing trust in synthetic data is zero:
\[
\mathbb{E}_{\pi_{sr,\lambda^*}}\!\left[\log L(D_s^*(q)\mid\theta)\right]
=
\mathbb{E}_{\rho(\cdot\mid D_s^*(q),\lambda^*)}\!\left[\log L(D_s^*(q)\mid\theta)\right].
\]
At this $\lambda^*$, the influence of the synthetic data is aligned with the empirical evidence provided by real data. Calibration therefore usefully counteracts analyst subjectivity: prompt-induced variation persists, but only to the extent supported by real observations. Alternatively, one might also consider prompt calibration, where we adapt the prompt to match the synthetic data to real data. 

We will note that while the calibration route is feasible one still has to consider the implications that the choice of $\lambda$ implies. Recall that this parameter reflects the user's prior trust in the generative model. As such, the calibration exercise will need to be constrained to deliver values consistent with that interpretation. In particular, large values (close to $1$) of $\lambda$ would imply that the synthetic data be treated on par with real data. In the limit, the results would be based completely on synthetic data which would swamp real data, obviating the need for real data altogether. This creates the need to "shave" or constrain $\lambda$ to more reasonable levels. In what follows, we will assume that the user has, by approprately chosen methods, specified $\lambda$ based on their overall trust of the generative model and the perceived fidelity of the synthetic data to anticipated and/or calibration data.   

\subsection{Additional Strategies}

Beyond integration, conservative trust, and calibration, several practical tools can further mitigate subjectivity. Sensitivity analysis to controlled perturbations of $q$ reveals the dimensions of prompt heterogeneity that most affect inference. Domain constraints may be used to filter out synthetic datasets inconsistent with known structure. Adversarial or contrastive prompting can detect brittle or hallucinatory behavior. Finally, documenting prompts, model versions, temperatures, and seeds makes subjective choices transparent and reproducible. Together, these strategies ensure that the foundation prior reflects human heterogeneity in a principled and controlled fashion while avoiding undue dependence on any single subjective specification.

\section{Posterior Beliefs}

We now turn to the issue of using generative data to construct posterior beliefs about $\theta$. Since we now have the foundation prior available, we can incorporate that into a coherent Bayes framework straightforwardly. We write the posterior for $\theta$ conditional on some real data $D_r$ (and the generated synthetic data $D^*_s$) as
\begin{equation}
   \pi(\theta|D_r,D_s,\lambda) \propto L(D_r \mid \theta)L(D_s^* \mid \theta)^\lambda \pi(\theta)
\end{equation}
The key element in the above equation is again $\lambda$. Recall our discussion that this parameter reflects the prior trust that the user has in the foundation model and the generative process. The choice of this parameter is of critical consequence for inference about $\theta$. First, choosing $\lambda=0$ is inefficient, since it throws out information that the foundation model contains about $\theta$. However, as argued earlier, setting $\lambda=1$ (or $\geq 1$) is conceptually troubling. Since we can generate an infinite amount of synthetic "data" setting $\lambda \geq 1$ would obviate the need for any real data. In a sense the prior would swamp the data which does not seem desirable except in some very particular situations.  

We suggest that user or analyst choose the trust parameter $\lambda$ based on three elements completeness, context and cost. First, the completeness of the foundation model including its ability to produce logically consistent and contextually relevant responses, lack of systematic biases, and the fidelity it exhibits to real data will all impact the construction of the trust parameter. Second, we recognize that there are contexts where the user might be willing to trust the foundation model more. For example, there will be no need to seek additional data if we ask the model to generate responses for basic arithmetic operations. Of course, this depends on the model and the earlier point about completeness. Context can also manifest itself in the nature of the task at had. If, for example, the purpose is simply exploration, prototyping, model design or any other objective where the inference aspect is less relevant then synthetic data seems like a reasonable avenue. Prediction is one such task  - where we may be less worried about any subjectivity or bias as long as prediction objectives are met. This argument is in line with those articulated in \cite{ludwig_large_2025}. More broadly, any application where the user's priors are used can now be enhanced with foundation priors. We discuss these in the next section. Finally, there are cases where real data is costly. This could be because real data is unavailable, difficult to gather, or simply monetarily expensive. Either way, the user can make well-reasoned trade-offs along the trust-cost curve.  

We also note here that there are alternate ways of incorporating a measure of trust in to the framework. While we have focused our attention on the choice of $\lambda$, an alternative approach would be to fix $\lambda=1$ and make a choice of weights on, or the number of synthetic observations we generate and use. For example, we can implement the unit information prior (\cite{kass_reference_1995}) by simply making sure that the weights on the synthetic data ($D_s$) be such that the effective sample size is equivalent to a single observation of real data. If the user trusts the generative process, they may increase the effective sample size accordingly. Alternatively, one can adjust the sample size itself ($N_s$) to increase or decrease the strength of the prior.  The choice of the mechanism to adopt will depend on the application at hand. 

\subsection{Discussion}

In the above, we have assumed that the user starts with priors, obtains synthetic data, and then uses real data to update their beliefs about $\theta$. The framework is \emph{coherent} in that the ordering of the steps do not change the posterior beliefs. For example, if the real data were available, then the user could use the usual Bayesian update to start $\pi(\theta|D_r)\propto L(D_r|\theta)\pi(\theta)$  and then obtain synthetic data. Fine-tuning (discussed above) is one such example on conditioning on available data but we could just as easily condition on other forms of information as well. 

We conclude the discussion of the foundation prior by noting that the exact formulation of the prior (\ref{fprior}) is an artifact of specific functional assumptions made. The structure can be generalized, for example, to work with moments (rather than likelihoods), theoretical conditions, or other forms of informational constraints. The key point remains that foundation models interact in complex ways with the user to generate informative but subjective outputs.  

\section{Use Cases and Applications}

We now take a broader perspective on where and how synthetic data generated via foundation models can be used in empirical settings. Any context that allows the user to use their subjective beliefs will offer an opportunity for the use of synthetic data and foundation priors. Such beliefs could be incorporated via synthetic data, via construction of latent constructs, or by specifying the foundation prior itself. In what follows, we will be philosophically agnostic (between Frequentist and Bayesian paradigms) and discuss applications and use cases of the generative data (and foundation priors) in analytic tasks.

Our standing assumption will be one of no leakage. In other words, any "real" data, if available, will not overlap the data used in the training of the foundation foundation model. We refer the reader to \cite{ludwig_large_2025} for a more detailed discussion on leakage related issues.    

\subsection{The Simple Stuff}

There are a number of simple tasks where the foundation prior or generative data might help or augment decisions made by researchers. For example, one could use synthetic data to compute initial values for $\theta$ that can then be used in an optimizer with real data. Typcially, these starting values would be drawn from the researcher's prior $\pi(\theta)$ which may be less informative than those generated by the foundation prior.  

Another application is that of power calculations and sample size calculations. A key ingredient in these procedure (Frequentist or Bayesian) is the effect size which is typically based on the user's prior beliefs. Using foundation models to generate data and calibrate effect sizes might allow for a more informative construction of these quantities and potentially more accurate outcomes. More generally, such data can be used to inform the design of experiments. For example, one could think of informative priors on the model space or on sparsity structures along with effect sizes to help in the construction of complicated research and experiment design \citep{arai_generative_2025}.  

\subsection{Less Simple Stuff}

The informativeness of the foundation priors can also be used to increase efficiency of estimators. Consider the task of learning a function using Gaussian Processes (GPs). We can refine Gaussian Process (GP) priors using synthetic data \( D_s \) before incorporating real data for function learning. The idea is simple, use generated data to \emph{narrow} the prior GP and then update with real data. The amount of synthetic data used $n_s$ controls the amount of prior information used. If done appropriately, the informativeness of foundation models could help in ensuring that learning from real data is more efficient and robust.

Another example is to augment models with the foundation priors where data is sparse or unavailable. For example, \cite{li_frontiers_2024}, show compelling evidence that human-generated pairwise brand similarity for automobiles can be approximated quite well by LLMs (agreement rates $>75\%$). One could use such generated data to construct substitution patterns to inform the covariance matrix of brand choice models. This would provide a less expensive path to obtaining realistic substitution pattens, than collecting stated preference data as done by \cite{berry_differentiated_2004}.       

\subsection{Hierarchical Models and Random Coefficients}

Recall that mixture models can be thought of as hierarchical models in a Bayesian sense. As such, the mixture distribution is simply a structured prior over some latent constructs. This interpretation allows us to inject foundation priors (or synthetic data) in lieu of standard parametric mixture distributions.

\subsubsection{Using Synthetic Data ($D_s$) for Mixture Models}

As an example, we propose a two-step estimation procedure that integrates synthetic data \( D_s \) generated from a foundation model into the \cite{bajari_linear_2007} (BFR) random coefficients estimator for discrete choice models. The estimator consists of:
\begin{enumerate}
\def\labelenumi{\alph{enumi}.}
\item \emph{Estimating the Basis Coefficients \( \beta_r \) Using Synthetic Data}
\begin{enumerate}
   \item Generate synthetic choice data \( D_s \) using a foundation model with an optimized prompt \( q^* \):
   \begin{equation}
   D_s = \{ (Y_s^{(i)}, X_r^{(i)}) \}_{i=1}^{N_s}, \quad Y_s^{(i)} \sim P_{\text{LLM}}(j | X_r^{(i)}, q^*).
   \end{equation}
   
   Here, \( P_{\text{LLM}}(j | X_r^{(i)}, q^*) \) represents the synthetic conditional choice probabilities implied by the foundation prior. Note that  $X_r^{(i)}$ are real data, while the $Y_s^{(i)}$ are synthetic choices generated by the model. Repeat this step $b=1\dots B$  times using a variety of prompts that are engineered to generate $Y_s^{(i,b)}$ such that the variation in the data covers the parameter space effectively. This procedure can be deliberately subjective.  

 \item  Estimate the basis coefficients \( \{ \beta^{(b)} \}_{b=1}^{B} \) by solving  
   \begin{equation}
   \hat{\beta}^{(b)} = \arg \max_{\beta} \sum_{i=1}^{N_s} \sum_{j=1}^{J} \mathbf{1}(Y_s^{(i,b))}=j) \ln P(Y=j|X_r^{(i)},\beta),
   \end{equation}
where the $P(Y=j|X_r^{(i)}$ are the choice probabilities based on the chosen structural model (say Multinomial Logit).  This step recovers estimated coefficients \( \hat{\beta}^{(b)} \) that potentially covers the parameter space better than uninformed prior would. The degree to which this coverage is better will depend on the information content of the language model as well as the subjective priors of the researcher. The researcher can augment the \( \hat{\beta}^{(b)} \) with additional elements if desired. 
\end{enumerate}

\item \emph{Estimating the Mixture Weights \( u_r \) Using Real Market Data}

Once the basis coefficients \( \{\hat{\beta}_{b}\}_{b=1}^{B} \) are obtained, the next step is to estimate the mixture weights \( \{ w_{b} \}_{b=1}^{B} \) using real market share data \( Y_{j,t} \) following exactly the steps in \cite{bajari_linear_2007}:
\begin{enumerate}
\item Define the market share equation:
   \begin{equation}
      Y_{r}^{(i)}
      =
      \sum_{b=1}^{B}
         w_{b}\,
         P_{r}\!\left(Y=j \mid X_{r}^{(i)}, \hat{\beta}_{b}\right)
      + \eta^{(i)},
   \end{equation}
   where \( P_{r}(Y=j \mid X_{r}^{(i)}, \hat{\beta}_{b}) \) is the predicted probability of choice for type \( b \), and \( \eta^{(i)} \) is an error term. Estimate the mixture weights \( \{ w_{b} \}_{b=1}^{B} \) using standard regression procedures. If the constraints \( \sum_{b=1}^{B} w_{b} = 1 \) and \( w_{b} \ge 0 \) are imposed, this becomes an inequality-constrained least squares (ICLS) problem.
\end{enumerate}

\end{enumerate}

The incorporation of the foundation prior into the framework does not alter the properties of the BFR estimator in any way. The example illustrates how the foundation prior can be used to inform the structure of the heterogeneity while ensuring that estimation remains grounded in real market data. The ideas can be used in the more sophisticated framework articulated in \cite{fox_simple_2011}. More generally, one can use the foundation prior as an informative prior for the mixing distribution. This has two advantages, it makes the heterogeneity more informed while also simplifying the estimation process.  

\subsubsection{Latent Constructs: Partially Linear Models with Synthetic Data \( D_s \)}

A different interpretation of random coefficients is via latent constructs. For example,  we might use foundation models to generate regressors (conditionally) to augment the covariate set. This is similar to ideas in \cite{goolsbee_consumer_2004} who draw from the empirical distribution of consumer demographics. One can replace the distribution of demographics with the foundation prior.

Below, we extend the framework of partially linear models by incorporating synthetic data \( D_s \) generated from a foundation model. The goal of the exercise is to estimate $\theta$ efficiently. The regression controls for covariates via the nuisance function $g$. It is well known that allowing for $g$ can significantly soak up noise in the regression thereby increasing precision in $\theta$. The framework is then:

\begin{equation}
    Y = T \theta + g(X) + U, \quad E[U | X, T] = 0.
\end{equation}

Our idea is to introduce additional covariates \( Z \sim P_{\text{LLM}}(Z | X) \) to improve the estimation of the structural function \( g(X) \) in the model. We can use the foundation model to synthesize additional covariates \( Z \sim P_{\text{LLM}}(Z | X) \) that contain auxiliary information about \( X \) and expand the feature space in \( g \). The nonparametric component is now augmented with $Z$ and we write the nuisance function as \( g(X,Z) \). Solving the expected mean squared error minimization problem with some smoothness penalty $\mathcal{P}$:
    \begin{equation}
    \{\hat{g},\hat{\theta}\} = \arg\min_{g,\theta}  \left[ \sum_{i=1}^{n} \mathbb{E}_{Z|X}(Y_i - T_i \theta - g(X_i, Z))^2 \right] + \mathcal{P}(g),
    \end{equation}
    gives us the quantities of interest. Note that the expectation is approximated using Monte Carlo integration with \( \{ Z^{(m)} \}_{m=1}^{M} \) drawn from \( P_{\text{LLM}}(Z | X) \). While the above is an illustrative setup, we can always extend it to more involved settings such as those outlined in \cite{chernozhukov_doubledebiased_2018}. Similar ideas are explored, from a Frequentist perspective, in \cite{angelopoulos_prediction-powered_2023} and \cite{zrnic_cross-prediction-powered_2024}. 

In the above examples, we have used the foundation prior to inform latent constructs that can be implicitly validated by real data via the estimation process. We contrast this with an approach that would more explicitly condition on the information contained in the foundation prior as "real" variation. For example, one could generate candidate draws of $\theta$ from the foundation prior and use that as the mixing distribution without any additional parameters. This implies a level of "trust" in the foundation model that may be unfounded. More generally, we advocate the use of the foundation prior (and synthetic data) to inform heterogeneity or other latent constructs with a mechanism that allows real data to reject them if needed.   

\subsection{Beliefs}

The foundation prior can also be used, generally speaking, as a stand-in for beliefs of latent agents. In ongoing work we use the foundation priors idea to synthesize data to match a variety of beliefs about the trajectory of the state space in a dynamic choice model. By exploring which beliefs (or data based on those beliefs) rationalize the real choices of economic agents we might be able to better understand the mechanism of dynamic choice.

The idea of generating beliefs has been explored in recent work by \cite{bybee_surveying_2023}. The paper outlines a procedure to generate economic expectations by applying large language models to historical news, producing measures that align with surveys and highlight behavioral biases. Using 120 years of expectations, the author shows that sentiment-driven industries have a higher crash probability and lower future returns, with evidence suggesting return extrapolation as a key mechanism in bubble formation. While the paper ignores the subjectivity of the responses and the prompt engineering process the idea of using foundation models to generate belief distributions is novel and worth pursuing. 

\section{Implications}

Viewing outputs from foundation models as components of a subjective prior, rather than as objective data, has several practical implications for empirical work and for how these systems are deployed in organizations and policy settings. The central message is simple: generative outputs can be useful, often highly so, but only when treated as structured prior information whose construction and influence are explicitly documented and controlled.

For empirical researchers, the framework suggests a different default practice for using synthetic data. Rather than treating model responses as if they were additional observations, they should be viewed as draws from the prior-predictive distribution implied by the foundation prior in (\ref{fprior}), with explicit dependence on the prompt $q$ and trust parameter $\lambda$. This perspective naturally leads to procedures that (i) register the prompts and prompt-engineering steps used to produce $D_s^*$, (ii) report the chosen value of $\lambda$ and its justification, and (iii) assess how posterior conclusions change under plausible alternatives for both. In many applications, this will shift synthetic data from the main object of estimation to a device for refining priors, designing experiments, constructing latent structures, or generating benchmarks for power and specification checks.

The same logic applies to industry and business settings where foundation models are increasingly embedded in decision workflows. When synthetic customer responses, counterfactual scenarios, or simulated behaviors are generated, they effectively define a prior over outcomes and states that reflect both the model’s training data and the organization’s anticipations. Using these as if they were measurements can lead to decisions that primarily reflect internal beliefs rather than external evidence. Treating them as priors instead encourages a more cautious design: synthetic outputs can be used to stress-test strategies, explore model space, or inform prior distributions in forecasting and risk analysis, but they should be calibrated against real data whenever available and downweighted when heterogeneity in prompting or model behavior is large.

The framework also has implications for agentic AI systems, where foundation models act as internal belief engines for agents that plan and act. In such systems, the agent’s beliefs about the world are, in effect, foundation priors: they summarize what the model “expects” conditional on prompts, context, and prior interactions. Small changes in prompts or interaction protocols can therefore induce meaningful changes in agent behavior. The tools developed here suggest that these internal beliefs should be treated as explicit design objects, with trust parameters, calibration steps, and sensitivity analyses defined in advance. Robust decision rules can then be framed as operating on belief distributions that are understood to be subjective, rather than as acting on purportedly objective state estimates.

Finally, the analysis underscores the continuing importance of real data. Because synthetic data can be generated at negligible cost and in effectively unlimited volume, there is a temptation to allow it to drive inference. The foundation prior perspective makes clear why this is problematic: if $\lambda$ is chosen too aggressively, synthetic information can swamp real observations, and the resulting conclusions become reflections of anticipatory beliefs rather than empirical constraints. The calibration ideas in the previous section provide some guidance to the choice of $\lambda$, although ultimatley an appropriate calibration requires adjustments for subjective trust. More generally, the value of real data in this framework is precisely that it can confirm, correct, or reject the subjective structure encoded in the foundation prior.

In sum, the implications of our analysis are operational rather than grand. Foundation models can and should be used as powerful prior generators and synthetic-data engines, but their outputs are best understood as subjective, prompt-dependent beliefs. Making those beliefs explicit, assigning them a controlled role through the trust parameter $\lambda$, and anchoring them whenever possible to real data provides a coherent path for incorporating foundation models into empirical analysis and decision-making without conflating synthetic information with the phenomena we ultimately care to learn about.

\section{Discussion and Conclusion}
In this paper, we introduced and formalized the idea of a “foundation prior,” which treats responses from large foundation models as arising from a potentially informative but inherently subjective prior. The central mechanism behind this inherent subjectivity is straightforward: generative outputs arise from an interaction between a user’s anticipations, the prompt $q$ they choose, and the internal representations of the foundation model. The prompt-engineering process, far from being a neutral technical step, functions as a selective filter that shapes which synthetic data are accepted and which are rejected. As a result, synthetic data inherit both the model’s learned structure and the user’s prior beliefs, often in ways that are opaque or unacknowledged.

Interpreting these outputs as components of a \emph{foundation prior} clarifies both their value and their limitations. They can be informative, sometimes highly so, and can guide estimation, experiment design, or model specification. But they should influence inference only through an explicitly parameterized trust weight $\lambda$ and never by being treated as if they were drawn from the same process as empirical observations. When framed this way, synthetic data become a powerful source of structured prior information rather than a surrogate for real evidence. The tools developed in this paper—integrating across heterogeneous prompts, tempering the influence of synthetic data through conservative trust, and calibrating their effect using real observations—offer a principled way to control this subjectivity.

The broader implication is that foundation models do not eliminate the need for empirical data; they heighten it. Real data serve as the anchor that disciplines the subjective structure embedded in the foundation prior. Without such anchoring, the iterative prompt process risks reinforcing the user’s prior expectations, producing a form of epistemic circularity. With anchoring, however, foundation priors can serve as an efficient and transparent way to inject domain knowledge, structure high-dimensional spaces, or help navigate problems where real data are scarce.

Ultimately, this paper reframes synthetic outputs not as answers from an oracle but as belief statements shaped jointly by the user and the model. Recognizing this subjectivity, and managing it explicitly, is essential for using foundation models responsibly in empirical research, decision-making, and downstream applications such as the design of agentic systems. The task ahead is not to chase objectivity in generative outputs, but to treat their subjectivity as a feature that must be modeled, documented, and calibrated—so that synthetic information and real data can interact in a coherent and mutually informative way. We anticipate that as foundation models grow more robust and ubiquitous, these tools will become a standard part of empirical workflows. The key will be responsible implementation and transparent documentation, ensuring that the inevitable subjectivity introduced by prompt-engineering and trust decisions do not obscure meaningful insights about the phenomena under study.

\newpage
\bibliographystyle{aea}
\bibliography{refs}{}

\end{document}